# What does ChatGPT return about human values? Exploring value bias in ChatGPT using a descriptive value theory

Ronald Fischer[1,2], Markus Luczak-Roesch[3] & Johannes A Karl[4,2]

## Abstract

There has been concern about ideological basis and possible discrimination in text generated by Large Language Models (LLMs). We test possible value biases in ChatGPT using a psychological value theory. We designed a simple experiment in which we used a number of different probes derived from Schwartz's basic value theory (items from the revised Portrait Value Questionnaire, the value type definitions, value names). We prompted ChatGPT via the OpenAI API repeatedly to generate text and then analyzed the generated corpus for value content with a theory-driven value dictionary using a bag of words approach. Overall, we found little evidence of explicit value bias. The results showed sufficient construct and discriminant validity for the generated text in line with the theoretical predictions of the psychological model, which suggests that the value content was carried through into the outputs with high fidelity. We saw some merging of socially oriented values, which may suggest that these values are less clearly differentiated at a linguistic level or alternatively, this mixing may reflect underlying universal human motivations. We outline some possible applications of our findings for both applications of ChatGPT for corporate usage and policy making as well as future research avenues. We also highlight possible implications of this relatively high-fidelity replication of motivational content using a linguistic model for the theorizing about human values.

[1]Neuroscience and Neuroinformatics Research Group, Institute D'Or for Research and Teaching, Rua Diniz Cordeiro, 30 - Botafogo, Rio de Janeiro - RJ, 22281-100, Brazil; ronald.fischer@idor.org
[2]School of Psychology, EA 314, Easterfield Building, Gate 3, Kelburn Parade, Wellington, 6012, New Zealand
[3]School of Information Management, RH 410, Rutherford House, 23 Lambton Quay, Wellington, 6011, New Zealand; markus.luczak-roesch@vuw.ac
[4]School of Psychology, Dublin City University, Dublin, Ireland; Johannes.karl@dcu.ie; Johannes.a.karl@gmail.com

**Code & data**: https://osf.io/w46nq/

**Funding**: None.

**Acknowledgements**: We would like to thank Larissa Hartle and Joanne Sneddon as well as participants of the April 4th, 2023 Value Seminar at the Centre for Human and Cultural Values, University of Western Australia for valuable feedback and suggestions.

Throughout history, texts have been authored by individuals or groups of individuals sharing collective values. With the emergence of a new generation of Large Language Models (LLMs), systems powered by artificial intelligence (AI) have started to generate text that carries value content, which is used in public messaging, marketing, and communication across all levels of society (Chia, 2023; Constantz, 2023; Roose, 2023). At the same time, there is increasing concern that these LLMs may not express the values of their creators or the training data that were used to develop the models (Bowman, 2023) and there is increasing evidence that they produce biased or toxic output (Bender & Koller, 2020; Chan et al., 2023; Pellert et al., 2022; Rozado, 2023).

Values are organized sets of beliefs that are used to evaluate actors, actions, and events, guiding the behaviors of individuals and collectives. Not surprisingly, values are at the core of public life, central to evidence-based policy making, and communicated in vision and mission statements of modern institutions. For individuals, values underlie voting patterns, support for public policies and diversity initiatives, patterns of cooperation, are used to justify discrimination and violence, and influence their wellbeing (Boer & Fischer, 2013; Caprara et al., 2017; Fischer & Karl, 2022; Maio, 2017; Sagiv et al., 2011; Sagiv & Schwartz, 2022; Scharfbillig et al., 2021).

One of the most pressing issues with the widespread adoption of LLM-based text generation is whether these tools introduce or alter expected value content, considering that there there these models become increasingly autonomous and unpredictable, and that there are no reliable techniques available to steer the behavior of LLMs (Bowman, 2023). Bias and discrimination in online and offline behavior is motivated by values, making the detection of possible value biases and systematic tests in these applications of utmost importance (Chan et al., 2023). Similarly, embedded value differences in pre-trained language across languages and cultures might be likely, but these continue to be poorly explored and documented (Arora et al., 2022). At the same time, available well-established descriptive theories of values within psychology (Sagiv & Schwartz, 2022; Schwartz et al., 2012) have significant potential to allow for a systematic test whether LLMs do produce bias and if so in what direction. If we were to feed value-theory driven prompts to LLMs, would the text generated by them adhere to the value content which the statement aims to elicit, or can we detect bias shifts towards specific value content? If, for example, the value content of an LLM output diverges from theoretically intended domains this might lead to inadvertent priming effects that prime unintended salient values which were not intended. Such a shift in value content could result in outcomes that undermine intended messages and run counter to the intended purpose.

In the current study we focus on ChatGPT as a specific instance of an LLM, which has received significant attention in the media for its ability to generate convincing human-like texts. ChatGPT is a chatbot application that leverages a version of a GPT (Generative Pre-trained Transformer) LLM. The GPT models are designed to generate text by predicting the most likely next word in a sequence, based on a large amount of existing text data that they were trained on. One of the key features of ChatGPT is its ability to understand and respond to natural language instructions (so called prompts), which makes it easier and more intuitive for users to interact with it. The purpose of ChatGPT is to provide an AI-powered natural language agent that can engage in a conversational style with users. Nevertheless, while this is the original purpose, ChatGPT has been started to be used as tool to write and research a wide range of texts such as newspaper articles, policy texts and speeches with unclear consequences, if there are currently unknown bias shifts in ChatGPT's outputs.

Our study provides baseline data on the ability of ChatGPT to generate theory-driven value-consistent content, aiming to provide insight into potential biases and pitfalls users and developers of these applications should be made aware of. We focus on the text generated in response to theory-driven prompts into ChatGPT via the professional OpenAI application programming interface (API) and evaluate to what extent the generated text includes words that have been included in a previously validated theory-driven value dictionary. Dictionary or word list methods have the distinct advantage that they are straightforward, simple and transparent. If a certain word occurs within a generated text, it implies that the concept has been invoked (Tausczik & Pennebaker, 2010). We also provide initial baseline information on the performance of the dictionary used and variations of theory-based prompts.

Our approach is guided by the most comprehensive psychological theory of values available (Fischer, 2017; Sagiv & Schwartz, 2022; Schwartz et al., 2012). The theory suggests that values form a circular motivational continuum, where the motivations that values express blend into one another, much like the colors in a color circle. This circular continuum can be divided into wedges, with the number of wedges depending on the research goal and the desired precision of measurement.

Figure 1. Schematic presentation of the basic value theory

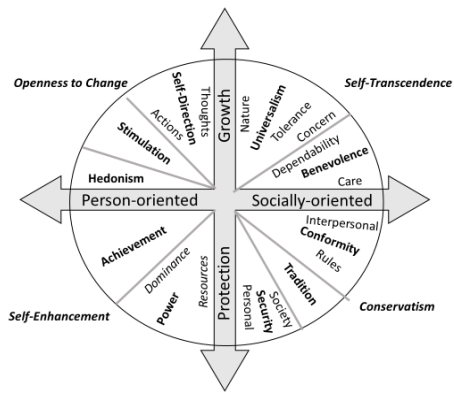

Figure 1 shows examples of the circle divided into as few as two broadly defined value types or as many as 19 more narrowly defined value types. These values can further be grouped using various sets of two dimensions, the most common sets of dimensions include (a) openness to change versus conservatism values and self-enhancement versus self-transcendence values, (b) personal focus versus social focus values, and (c) growth versus self-protection values. It is important to note that, although the circle in Figure 1 locates the values as equidistant, the theory only refers to their order. Table 1 presents definitions of the 4, 10, and 19 value types as they are most often used in the literature.

This circular continuum of values captures three essential features of relations among values. Firstly, adjacent value types in the circle are motivationally compatible and can be pursued in the same action. For example, the value types of hedonism and stimulation are adjacent and compatible. Secondly, value types located on opposing sides of the circle are motivationally opposed and typically cannot be pursued in the same action (e.g., stimulation and security). Finally, the motivational compatibility between value types decreases with the distance between them around the circle. Overall, Schwartz's theory provides a useful framework for understanding how values relate to each other and how they can influence behavior. It highlights the complexity of human motivations and the need to consider multiple values when trying to understand why people act or do not act in certain ways.

Several different instruments have been developed over the years that allow differentiating the value types to different degrees. The revised version of the Portrait Value Questionnaire is the most recently developed tool separating the 19 value types, which offers new insights into the relations of values to attitudes, behaviors, personality, and demographics (Schwartz et al., 2012; Schwartz & Cieciuch, 2022). To capture values expressed in written text, Ponizovskiy and colleagues (2020) developed a theory-driven personal values dictionary, which assesses individual value priorities by analyzing references to values in natural language. Their approach was based on the idea that authors of a text refer to values they consider important more frequently in their language (Bardi et al., 2008; Boyd & Pennebaker, 2017; Tausczik & Pennebaker, 2010). The central idea is that values are expressed through the choice of words that invoke a particular value when writing about an object, event, state of being, an evaluation, or other types of descriptions. For example, the value type of security may be invoked by using 'alarm', 'caution', 'danger' or 'threat'. The final list of 1068 dictionary terms was derived from a text corpora generated by more than 180,000 people and considering internal and external validity criteria (typically using demographic, self-report and other conceptually related dictionary scores) (Ponizovskiy et al., 2020). This dictionary has been used to score media content, online text and life history narratives (Amozegar, 2021; Fischer et al., 2022; Gutierrez, 2021; Lanning et al., 2021; Ponizovskiy et al., 2020; Schwartz, 1992).

**Our study**
We assumed that an LLM-based chatbot can be used to generate textual content that maps onto a descriptive theory of human values (Schwartz, 1992), and set up a simple experiment to test this assumption in which we used a number of different probes derived from Schwartz's theory and current survey measures of values. We prompted ChatGPT via the OpenAI API repeatedly to generate text and then analyzed the generated corpus for value content with the theory-driven value dictionary (Ponizovskiy et al., 2020). By comparing it with the theory-driven dictionary, we can quantitatively evaluate the extent to which ChatGPT maintains the value orientation of the prompt when generating novel text or whether there are systematic bias shifts. We use a bag of words approach together with the expected frequency patterns specified by the psychological theory to examine in what direction shifts may be occurring. Considering the questions of possible ideological bias, we can also use previous research on the links between political and social attitudes and this value theory (Boer & Fischer, 2013). A higher frequency and overrepresentation of values that are growth-oriented (universalism, benevolence, self-direction and stimulation in particular) would suggest a shift towards a liberal position, whereas higher frequency and overrepresentation of values that are self-protective (conformity, tradition, security and power in particular) would suggest a shift towards a conservative position.

Table 1. Overview of the value types include in the basic
theory of values

| Value Type | Definition | Fine-grained Value Types | Motivational focus |
|---|---|---|---|
| Benevolence | Preoccupation with the wellbeing of individuals emotionally close to the person | Benevolence—Dependability: Being a reliable and trustworthy member of the ingroup<br>Benevolence—Caring: Devotion to the welfare of ingroup members | socially-oriented; growth-oriented |
| Universalism | Appreciation, tolerance and protection of the wellbeing of all people and nature, independent of the emotional closeness | Universalism—Concern: Commitment to equality, justice and protection for all people<br>Universalism—Nature: Preservation of the natural environment<br>Universalism—Tolerance: Acceptance and understanding of those who are different from oneself | socially-oriented; growth-oriented |
| Conformity | Restraint of actions or impulses that may upset others or violate social norms and expectations | Conformity—Rules: Compliance with rules, laws, and formal obligations<br>Conformity—Interpersonal: Avoidance of upsetting or harming other people | socially-oriented; self-protective |
| Tradition | Preservation, respect and commitment to norms and customs of one's traditional culture or religion | Tradition (as defined) | socially-oriented; self-protective |
| Security | Concern with the safety and stability of the self, close others as well as society | Security—Personal: Safety in one's immediate environment<br>Security—Societal: Safety and stability in the wider society | socially-oriented; self-protective |
| Power | Control and dominance of other people and resources | Power—Dominance: Power through exercising control over people<br>Power—Resources: Power through control of material and social resources | person-focused; self-protective |
| Achievement | Demonstrating success and competence according to socially accepted standards | Achievement (as defined) | person-focused; self-protective |
| Hedonism | Striving for pleasure and gratification of personal desires | Hedonism (as defined) | person-focused; growth-oriented |
| Stimulation | Seeking excitement, challenges and novelty in life | Stimulation (as defined) | person-focused; growth-oriented |
| Self-Direction | Independence in both thought and action | Self-Direction—Thought: Freedom to cultivate one's own ideas and abilities;<br>Self-Direction—Action Freedom to determine one's own actions | person-focused; growth-oriented |

## Method

ChatGPT text generation procedure
The focus of our analysis on content that ChatGPT generates when prompted to elaborate on input based on the basic theory of human values (survey items, definitions, value names). To minimize the linguistic bias added by our prompt beyond the actual theory content, we used the following very simple instructional prompt template:

> Prompt template: "Elaborate on: [item text]"
> Example prompt: "Elaborate on: It is important to him/her to form his/her views independently."

We ran this prompt for each item text 5 times, which resulted in 285 ChatGPT responses (57*5 runs).

We requested 300 tokens from the gpt-3.5-turbo model with the default parameter options *temperature=1* and *top_p=1*. As per API documentation this should allow for a moderate stochasticity in the generated responses (randomness can be decreased by lowering either the temperature or the top_p parameter to a minimum of 0, and by increasing the temperature parameter up to a maximum value of 2).

To control for further linguistic bias that may stem from the natural language characteristics of the individual items of the revised PVQ-57R, we executed two control runs (again requesting 300 response tokens). First, we used short definitions for the revised 19 value type from the same article, resulting in 95 ChatGPT responses (19*5 runs). Second, we used only the names of the 19 dimensions to rigorously minimize any additional linguistic overhead in the prompt (resulting in another 95 ChatGPT responses).

Occasionally ChatGPT answers with a variations of the pre-text "As an AI language model,…". We evaluated all retrieved variations of this pre-text and then removed the first sentence from all responses where that sentence matched the regular expression pattern "*AI language model,|As AI,|As a sentient AI|language model AI|As an AI*". Our tests indicated that there was no systematic pattern across inputs for this type of pre-text to occur.

Finally, we used quanteda (Benoit et al., 2022) to generate three document-term matrices from the three different response corpora, which we then filtered down to retain only terms that occur in the value dictionary. These matrices were the basis for all subsequent analysis steps.

## Value dictionary

We used the refined value dictionary developed by Ponizovskiy et al. (2020), which contains theory-driven selections of adjectives, verbs and nouns that have been shown to correlate with value priorities across a range of textual sources. The refined dictionary contains 1068 entries. The values are not equally represented in the dictionary: the smallest word set captures Security values (85 words) and the largest word set is the Self-direction value list (140 words; see Ponizovskiy et al., 2020). We counted the number of times each word in the dictionary appeared in the produced text. Following previous tests with this dictionary (Fischer et al., 2022), we used the raw scores without adjustment for text length.

## Calculations of bias effects

We are using a bag of words approach, counting the number of words that are included in each dictionary category in the text generated. The frequency count can then be compared to theoretically expected patterns. We derive a number of different metrics.

**Hits.** For each prompt (lines in Figure 2), we calculated how many words from the dictionary categories appeared in the generated text. We considered a score as a 'hit' when the highest score for the dictionary category was observed for a theoretically congruent value, e.g., when creating text using the security-personal prompt, we count a hit if the highest value dictionary score is observed for the security value dictionary score (which corresponds the black entries along the main diagonal in Figure 2).

**Validity benchmarks.** We created two scores which we term *concept validity* and *discriminant validity* score. First, the concept validity score captures the extent to which the conceptually expected dictionary score was highest relative to all other dictionary scores for a specific prompt. For this we calculated the ratio of the score of the value congruent score (security dictionary score for the security-personal item prompts, true positive) divided by all dictionary scores for that particular text corpus (as it contains all the true positives and false negatives). In Figure 2, this corresponds to the entries in each line (horizontally), we consider the theory-matching cell against the total number of words that can be matched to the dictionary for each prompt. This could also be seen as an indicator of sensitivity as it expresses the true positive scores against the combined true positives and false negatives. This concept validity benchmark results in a ratio between 0 and 1. A value of 1 would indicate that the created text by ChatGPT would feature only value-terms that are from the value dimension used in the prompt. A value of 0 would indicate that no theory-congruent term was included. As there are 10 options, and if we assume random distribution of scores across all dimensions, we can interpret any number above 0.10 as indicating reasonable sensitivity. In summary, similar to a construct validity coefficient this score captures the relative ability to correctly identify the value that was used to create the text.

Second, we calculated the equivalent score for each dictionary category (e.g., security) across all theory-driven prompts. We call this the discriminant validity benchmark, because it captures the extent to which the highest dictionary score is observed for the value-congruent prompt. In Figure 2, this corresponds to calculating the theory congruent cell (e.g., security dictionary count for the security-personal prompt for example) across the security dictionary counts for all prompts. By running this analysis we can examine whether ChatGPT did generate content aligned with the correct value dimension per prompt.

**Signal to noise ratio.** We calculated a signal to noise ratio as the number of hits to all false negatives. The same two types of signal to noise ratios as for the validity benchmarks can be computed (i.e. in horizontal and vertical direction). A value of 1 indicates that an equal number of hits and misses for the expected dimension were found. A value smaller than 1 indicates that more value terms were found in other texts compared to the value congruent text. A value larger 1 indicates that more theory-expected terms were identified in a text generated by a theory congruent prompt compared to any other prompt. This is the equivalent to the concept validity benchmark (calculating horizontally across all dictionary categories within a single prompt). Similarly, the same score could be calculated within each value dimension in the dictionary across all text generated by ChatGPT (vertical calculation in Figure 2). This is the discriminant validity benchmark equivalent.

**Theoretical profile match**. Since the theory specifies a circumplex pattern, we may expect some bleeding of terms within text. For example, when running a Security values prompt (see Figure 2), terms from theoretically adjacent values such as Tradition or Power values may also be more likely to be included in the generated text (see the top of Figure 2 for an expected frequency count for security dictionary words when using the security-personal prompt). As the values become more motivationally distant, we could expect a lower probability that ChatGPT will include those in the text (e.g., we would expect no mentioning of the motivationally opposing self-direction value words when using a security value prompt, see Figure 2). We therefore computed an expected pattern for each value and examined the detected value pattern within each theoretical prompt class.

**Underlying value structure in the text output.** We explored whether the dictionary identified value terms within each set of prompts would replicate the expected theoretical structure described by Schwartz (2002). Are motivationally related values more likely to show a theoretically expected blending pattern of hits in the generated text? To test this, we first converted the frequency matrix of dictionary counts per prompt into a correlation matrix (using Pearson rank order correlations). This correlation matrix was then transformed into a dissimilarity matrix and the resulting Euclidean distances were analyzed with an ordinal MDS specifying 2 dimensions. To examine whether the observed two-dimensional representation was similar to the theoretically expected model, we rotated the observed MDS solution to maximal similarity with the theoretical structure using the coordinates (Bilsky et al., 2011). We examined the overall similarity of the two dimensional structure with Tucker's phi and alienation coefficient as well as the similarity of the individual dimensions with Tucker's phi (Borg & Mair, 2022; Fischer & Fontaine, 2010). Values above .85 are typically seen as sufficient for replicating an expected structure (Fischer & Karl, 2019).

**Comparison variables.** We used the mean rating for the Schwartz Value Survey from representative or nearly representative samples collected in the early to mid-1990s. Data from the following samples was available: Australia, China, Finland, France, Germany, Israel, Netherlands, Russia, and South Africa.

To estimate the overall frequency of entries in the dictionary in the English language, we extracted the frequency of each dictionary term from the kaggle data set, which lists the counts of the 333,333 most commonly-used single words on the English language web, as derived from the Google Web Trillion Word Corpus (Tatman, 2017). All entries in the dictionary were included in this list of most-commonly used single words.

We will use these mean ratings and baseline frequencies of value-related words in English to examine whether the overall word counts are more aligned with mean preferences in representative samples or are due to word frequency effects in English (e.g., independent of value content).

**Supplementary analyses**
Finally, we ran baseline analyses to examine the dictionary scores for the PVQ-57R that we used as prompts. We did this because it is important to analyze how many dictionary terms are identified in the PVQ items themselves, as this may linguistically bias the output text generated.

Figure 2.
Schematic display of calculated bias effects

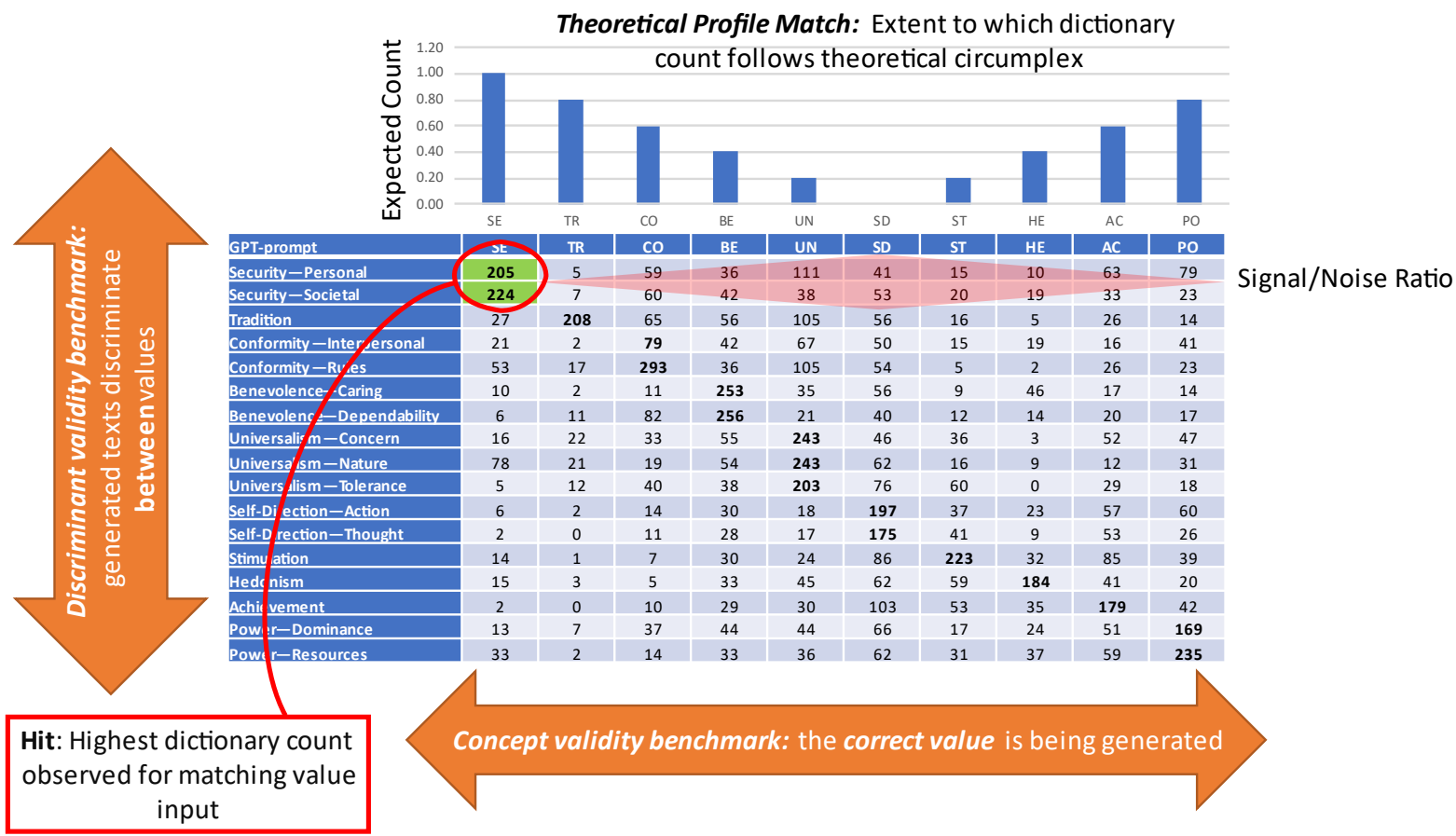

Note: The cells show the count of observed terms within each dictionary category per input prompt.

## Results

**Word frequency effects.** Figure 3 shows the most frequent terms per prompt. Using the PVQ-items as prompts, the most common terms in this corpus were lead (Power values), society (universalism values) and individual (self-direction values). The largest number of words related to Universalism values (1385), Self-Direction values (1285) and Benevolence (1095) in the dictionary. The fewest words in the generated corpus could be matched to Tradition (322), Hedonism (471) and Stimulation (665).

Using the definitions as prompts, by far the most frequent terms were society (universalism values), social (universalism), security (security values) and environment (universalism). Across this corpus, the most frequent value dimension was Universalism (906), followed by Self-Direction (443) and Conformity (382), the least frequent values were Tradition (131) and Hedonism (176) and Stimulation (218).

Finally, using value names as prompt input, the most frequent terms were social (universalism values), security (security values), society (universalism values) and power (power values). Focusing on the most frequent value terms from the dictionary identified in this corpus, the most frequent terms were from Universalism (639), followed by Self-Direction (519) and Power values (392). The least frequent values were Tradition (125), Hedonism (127) and Stimulation values (153). Considering the overall distribution mixing both progressive-liberal and conservative values in the most and least common values, we did not detect an explicit value bias in the overall distribution of the responses.

We then predicted the term frequency from both the median frequency of value terms in the dictionary in English and the mean rating of each value by representative samples. As can be seen in Table 3, overall the mean rating of the values is a relatively stronger predictor in our corpus compared to the frequency of words in English. This suggested that more highly rated values, controlling for the overall frequency of the terms in the dictionary, are more likely to be identified in this text corpus.

## Hits

The most straightforward benchmark for assessing the performance of ChatGPT is the number of hits, that it is whether the highest number of dictionary terms within a dimension were observed for the text generated by the corresponding value prompt. Examining hits across each prompt (construct validity), for the PVQ-57R prompts, we observed a 100% hit rate. In other words, the highest score per dimension was observed for text that was generated with a matching prompt for all values. For the definition prompts, hits were observed for all values, except for Benevolence Caring (highest frequency for Universalism dictionary values). For the value name prompts, all values showed hits for the corresponding chatGPT text except, Conformity-Interpersonal (highest score was observed for Universalism values), Stimulation values (highest frequency was observed for the conceptually adjacent Self-Direction values), and Universalism-concern (highest frequency for the adjacent Benevolence values).

When examining hits across value prompts within each dictionary value category (discriminant validity), we observed again 100% hits for the PVQ-57R prompts. For the value definitions, all prompts showed the highest dictionary score for the congruent value, except for Conformity – interpersonal (prompts using the Tradition value definition showed the second highest hit rate of 46 matching terms). For the value name prompts, we again observed the expected hits for all prompts except for Universalism – concern and Universalism – Nature (Security – Personal, Conformity – Rules & Conformity – Interpersonal prompts created text that featured a relatively large number of Universalism dictionary terms).

Figure 3a
Frequency distribution of dictionary counts in the PVQ57-R item condition

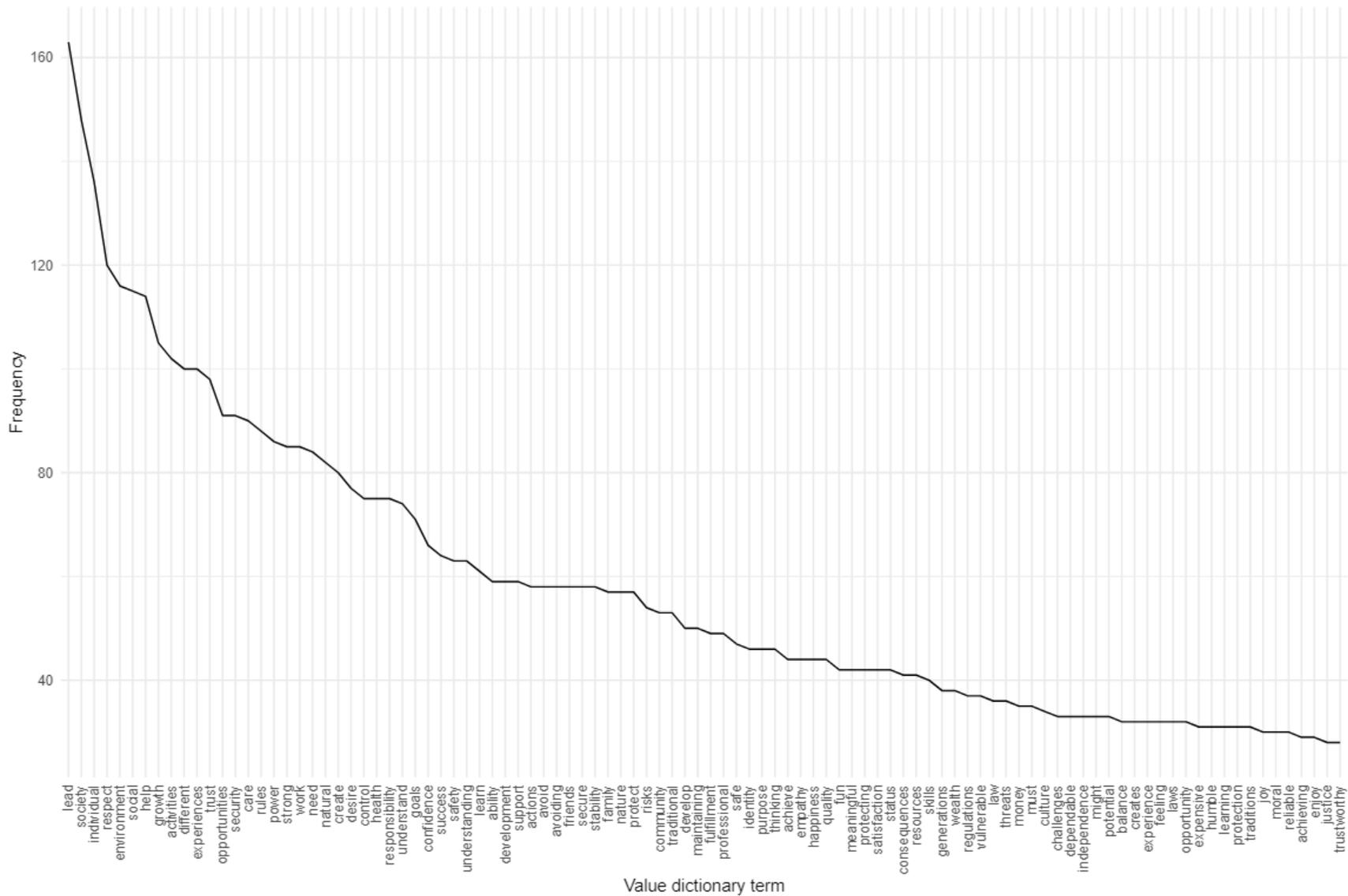

Figure 3b
Frequency distribution of dictionary counts in the value definition condition

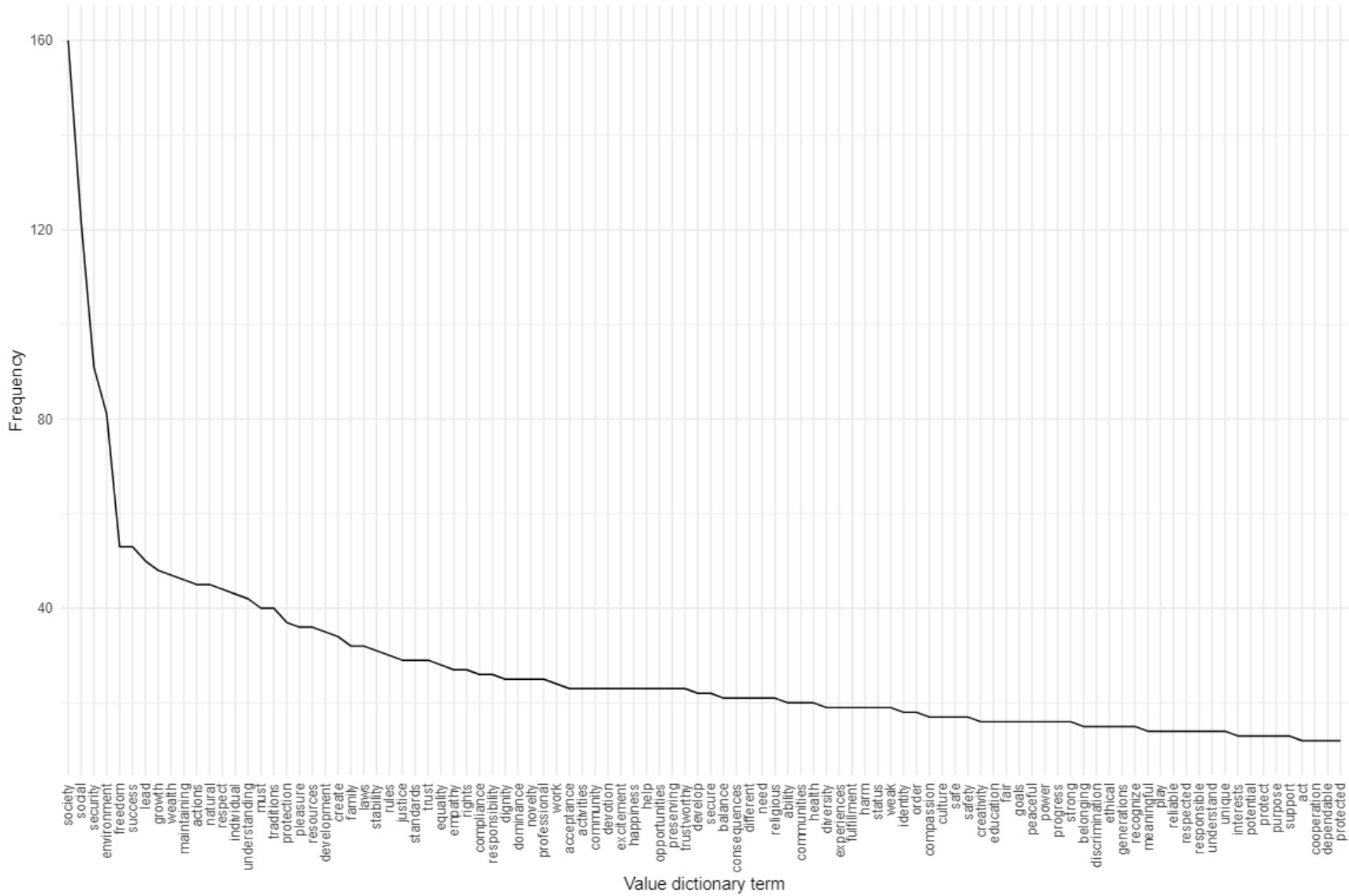

Figure 3c
Frequency distribution of dictionary counts in the value name condition

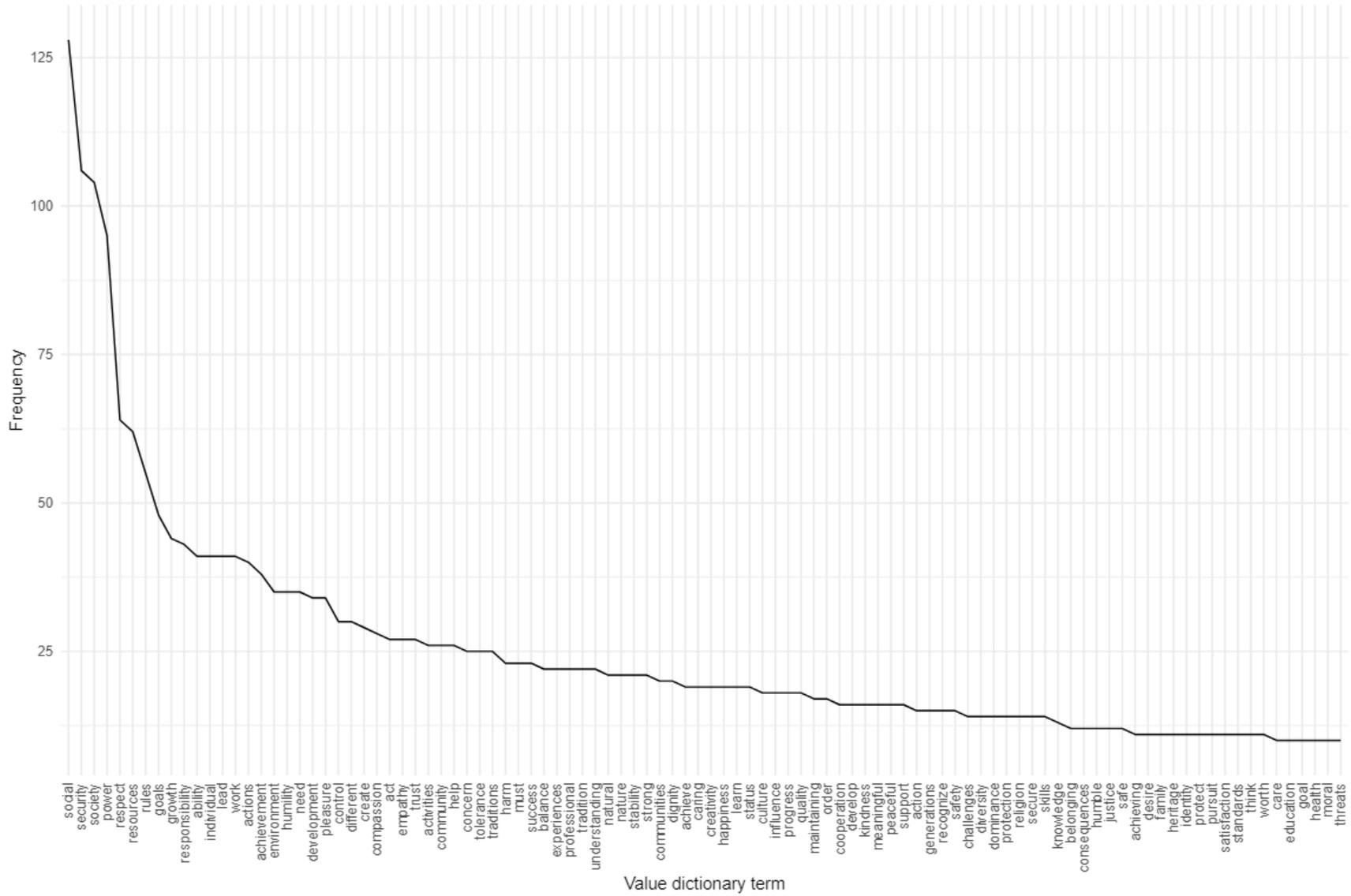

Table 3.
Summary of frequency and mean rating effects on dictionary counts

| Predictor | $b$ | $b$ 95% CI [LL, UL] | beta | beta 95% CI [LL, UL] | $sr^2$ | $sr^2$ 95% CI [LL, UL] | $r$ |
|---|---|---|---|---|---|---|---|
| **PVQ-57R** | | | | | | | |
| (Intercept) | 850.90** | [672.35, 1029.45] | | | | | |
| Word frequency | 180.97 | [-8.56, 370.50] | 0.54 | [-0.03, 1.11] | .29 | [-.12, .69] | .47 |
| Mean preference | 209.97* | [20.44, 399.50] | 0.63 | [0.06, 1.19] | .39 | [-.05, .83] | .56 |
| *Fit: R² = .605\*, 95% CI[.00,.76]* | | | | | | | |
| | | | | | | | |
| **Definition** | | | | | | | |
| (Intercept) | 144.50** | [78.61, 210.39] | | | | | |
| Word frequency | 20.57 | [-49.37, 90.52] | 0.24 | [-0.57, 1.05] | .06 | [-.20, .31] | .19 |
| Mean preference | 33.86 | [-36.09, 103.80] | 0.39 | [-0.42, 1.20] | .15 | [-.25, .55] | .36 |
| *Fit: R2 = .189, 95% CI[.00,.48]* | | | | | | | |
| | | | | | | | |
| **Value names** | | | | | | | |
| (Intercept) | 137.50** | [92.20, 182.80] | | | | | |
| Word frequency | 42.99 | [-5.09, 91.07] | 0.62 | [-0.07, 1.32] | .38 | [-.09, .86] | .60 |
| Mean preference | 13.12 | [-34.96, 61.21] | 0.19 | [-0.51, 0.89] | .04 | [-.14, .22] | .12 |
| *Fit: R² = .398, 95% CI[.00,.64]* | | | | | | | |

*Note.* A significant *b*-weight indicates the beta-weight and semi-partial correlation are also significant. *b* represents unstandardized regression weights. *beta* indicates the standardized regression weights. *sr²* represents the semi-partial correlation squared. *r* represents the zero-order correlation. *LL* and *UL* indicate the lower and upper limits of a confidence interval, respectively. All predictors were scaled.
* indicates *p* < .05. ** indicates *p* < .01.

## Validity benchmarks

Focusing first on the concept validity benchmark, which is the extent to which the correct value is generated by the model (horizontal calculation in Figure 2) we found relatively high validity coefficients. Considering randomly distributed data across the ten value types, we could expect a validity coefficient of .10. The mean was .417 for the PVQ-57R prompts, .414 for the definitions and .426 for the value names. The maximum validity coefficient was observed for the Power – dominance value name prompt (.58), the minimum was observed for the Conformity – interpersonal prompts using both the PVQ-57R and the definition as prompt input (.22). Therefore, the observed scores were well above the random distribution threshold of .10 and generally showed reasonable levels of concept validity (see Table 4).

Focusing on the discriminant validity benchmarks (vertical calculation in Figure 2), the average discriminant validity was .43, .42 and .45 for the PVQ-57R, definition and value name prompts, respectively (see Table 5). The highest discriminant validity was observed for Security values when using the value names as prompt input (.74). Overall, the consistently best discriminant performance across all prompts was observed for Tradition values. It is noteworthy that Tradition values were typically one of the lowest observed values in the corpus, but they appear to discriminate quite well across prompts.

**Signal to noise ratio.**

Focusing first on the construct validity signal to noise ratios, the average was .75, .75 and .78 for the three different prompts. As the signal to noise ratio can vary from 0 to infinity and are centered around 1 indicating an equal number of hits to misses, our averages indicate that hits on average had lower counts than all other value categories in the dictionary. The highest signal to noise ratio was observed for Power – Dominance when using the value name prompts. The lowest signal to noise ratio was observed for Conformity – interpersonal (.29 for both the PVQ-57R and definition prompts).

For the discriminant validity signal to noise ratio, the average ratios were .86, .80 and 1.02, respectively. This suggests that the discrimination across prompts was slightly better, especially when using value names as input. The highest signal to noise ratio was observed for Security values, which consistently showed relatively favorable discriminatory values (1.43, 1.20 and 2.79 for the PVQ-57R, definition and value name prompts, respectively). Achievement values showed the relative weakest signal to noise ratio here, with the values being .28, .28 and .45 for the PVQ-57R, definition and key word prompts.

**Pattern matching.**

This analysis focuses on the relative pattern of hits per type of prompt. Focusing on the concept validity comparisons, the average correlation between expected pattern and the observed pattern of word frequencies was .58, .48 and .52 for the PVQ-57R, definition and key word prompts, respectively. This analysis nevertheless has to be considered with caution because the number of terms in the dictionary varies across value types. Nevertheless, this could be considered a conservative test of the pattern matching per prompt.

For discriminant validity comparisons, the average correlation was slightly higher: .61, .58 and .60 for the PVQ-57R, definition and key word prompts, respectively. This analysis was not affected by different term sets in the dictionary, and as could be expected by removing this noise the pattern was somewhat higher. The highest discriminant validity pattern on average was found for Benevolence (r = .71) and Conformity (r = .70), the lowest pattern on average was observed for Power (r = .45) and Achievement (r = .47).

Table 4
Concept validity benchmarks

| Value Type | Validity | | | Signal/ Noise | | | Profile Matching | | |
|---|---|---|---|---|---|---|---|---|---|
| | PVQ57-R items | Definitions | Value names | PVQ57-R items | Definitions | Value names | PVQ57-R items | Definitions | Value names |
| AC | 0.37 | 0.30 | 0.50 | 0.59 | 0.43 | 1.00 | 0.52 | 0.31 | 0.52 |
| BE—Caring | 0.56 | 0.38 | 0.49 | 1.27 | 0.6 | 0.98 | 0.55 | 0.46 | 0.66 |
| BE—Dependability | 0.53 | 0.40 | 0.38 | 1.15 | 0.66 | 0.62 | 0.66 | 0.55 | 0.61 |
| CO—Interpersonal | 0.22 | 0.22 | 0.34 | 0.29 | 0.29 | 0.51 | 0.47 | 0.44 | 0.16 |
| CO—Rules | 0.48 | 0.48 | 0.35 | 0.91 | 0.93 | 0.54 | 0.65 | 0.42 | 0.38 |
| HE | 0.39 | 0.45 | 0.38 | 0.65 | 0.81 | 0.63 | 0.74 | 0.64 | 0.55 |
| PO—Dominance | 0.36 | 0.37 | 0.58 | 0.56 | 0.59 | 1.39 | 0.35 | 0.19 | 0.46 |
| PO—Resources | 0.43 | 0.42 | 0.53 | 0.77 | 0.74 | 1.13 | 0.52 | 0.52 | 0.48 |
| SE—Personal | 0.33 | 0.36 | 0.43 | 0.49 | 0.56 | 0.75 | 0.45 | 0.16 | 0.26 |
| SE—Societal | 0.43 | 0.41 | 0.42 | 0.76 | 0.70 | 0.72 | 0.45 | 0.31 | 0.39 |
| SD—Action | 0.44 | 0.48 | 0.42 | 0.80 | 0.94 | 0.73 | 0.56 | 0.59 | 0.54 |
| SD—Thought | 0.48 | 0.39 | 0.47 | 0.94 | 0.65 | 0.90 | 0.64 | 0.71 | 0.58 |
| ST | 0.41 | 0.45 | 0.29 | 0.70 | 0.83 | 0.42 | 0.77 | 0.75 | 0.85 |
| TR | 0.36 | 0.31 | 0.43 | 0.56 | 0.44 | 0.76 | 0.57 | 0.50 | 0.39 |
| UN—Concern | 0.44 | 0.53 | 0.29 | 0.78 | 1.14 | 0.42 | 0.58 | 0.44 | 0.71 |
| UN—Nature | 0.45 | 0.54 | 0.45 | 0.80 | 1.18 | 0.80 | 0.57 | 0.50 | 0.62 |
| UN—Tolerance | 0.42 | 0.54 | 0.50 | 0.73 | 1.18 | 1.00 | 0.74 | 0.60 | 0.64 |

Note: UN = Universalism, BE = Benevolence, CO = Conformity, TR = Tradition, SE = Security, PO = Power, AC = Achievement, HE = Hedonism, ST = Stimulation, SD = Self direction

Table 5
Discriminant validity benchmarks

| Value Dimensions | Discriminant validity | | | Signal/Noise | | | Profile Match | | |
|---|---|---|---|---|---|---|---|---|---|
| | PVQ-items | Definition | Key values | PVQ-items | Definition | Key values | PVQ-items | Definition | Key values |
| SE | 0.59 | 0.55 | 0.74 | 1.43 | 1.20 | 2.79 | 0.55 | 0.61 | 0.59 |
| CO | 0.44 | 0.45 | 0.37 | 0.80 | 0.80 | 0.59 | 0.59 | 0.71 | 0.79 |
| TR | 0.65 | 0.63 | 0.62 | 1.82 | 1.67 | 1.66 | 0.77 | 0.56 | 0.52 |
| BE | 0.46 | 0.31 | 0.40 | 0.87 | 0.45 | 0.66 | 0.67 | 0.76 | 0.71 |
| UN | 0.50 | 0.40 | 0.33 | 0.99 | 0.65 | 0.49 | 0.50 | 0.47 | 0.41 |
| SD | 0.29 | 0.37 | 0.30 | 0.41 | 0.58 | 0.42 | 0.61 | 0.56 | 0.61 |
| ST | 0.34 | 0.44 | 0.31 | 0.50 | 0.79 | 0.46 | 0.78 | 0.66 | 0.73 |
| HE | 0.39 | 0.45 | 0.47 | 0.64 | 0.81 | 0.90 | 0.60 | 0.69 | 0.65 |
| AC | 0.22 | 0.22 | 0.31 | 0.28 | 0.28 | 0.45 | 0.60 | 0.39 | 0.43 |
| PO | 0.45 | 0.43 | 0.64 | 0.82 | 0.77 | 1.74 | 0.48 | 0.36 | 0.52 |

Note:   UN = Universalism, BE = Benevolence, CO = Conformity, TR = Tradition, SE = Security, PO = Power, AC = Achievement, HE = Hedonism, ST = Stimulation, SD = Self direction

**Structural replication of the value space.**

We ran a MDS followed by procrustes rotation to examine the extent to which the observed value frequencies in the generated text replicated the theoretically predicted structure (Bilsky et al., 2011). For the PVQ-57R prompts, the overall similarity was .918 (alienation = .398), dimension 1 showed a Tucker's phi of .84 and dimension 2 yielding .41. The major shift was the location of Universalism with Conformity and Security values and the switched position between Self-Direction and Hedonism values. Examining the relative structural organization of the values based on the definitions prompts, the overall similarity was acceptable (Tucker´s Phi = .933, alienation = .359), the first MDS dimension was well replicated (Tucker's phi = .92), whereas the second dimension continued showing problems (Tuckers phi = .43). The major shift was the placement of Universalism values at the negative pole of dimension 2, whereas theoretically, it would have been expected at the positive end of dimension 2. Finally for the value names as prompts, the overall similarity was again relatively acceptable (Tucker´s Phi = .929, alienation = .371), first dimension was again relatively replicated (Tucker's phi = .83), whereas dimension 2 was showing lower replicability (Tucker's phi = .53). The major shift was again the placement of Universalism values close to Security and Conformity values as well as the switch of Self-Direction and Hedonism values.

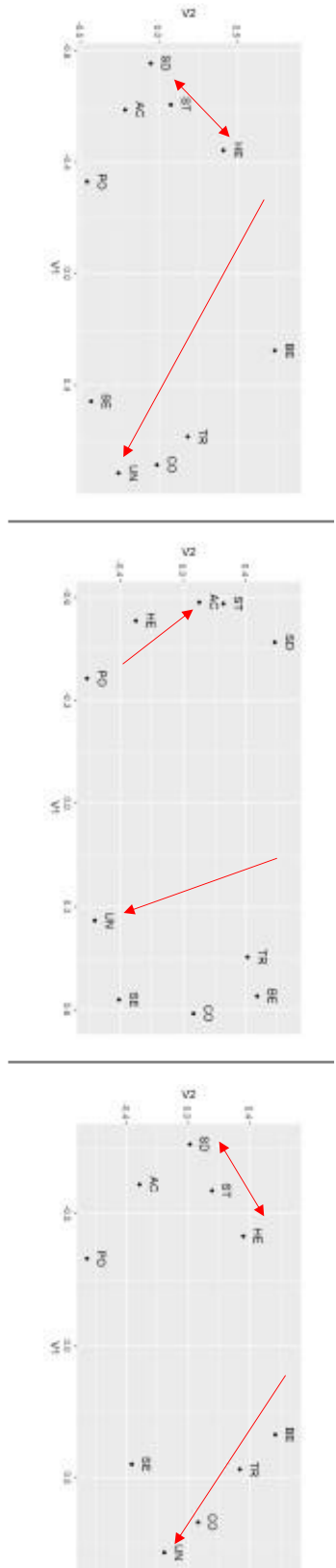

Figure 4

Relative internal structure of values rotated towards ideal structure, based on the observed dictionary frequencies in the corpus.

Note: Left panel: the PVQ57-R item prompts; Middle panel: the value definition prompts; Right panel: the value name prompts; UN = Universalism, BE = Benevolence, CO = Conformity, TR = Tradition, SE = Security, PO = Power, AC = Achievement, HE = Hedonism, ST = Stimulation, SD = Self direction

## Discussion

There has been much discussion on the prospects and limits of the current and future generations of LLMs. In our analysis we focus on one important aspect for social and political analyses, which is the extent to which the generated text maintains specific theory-guided value content embedded in the prompts. Our results suggest that overall, the ChatGPT version that we used shows relatively high fidelity in the generation of responses when compared to a theory-driven dictionary. This implies that specific value content used in user-defined prompts is carried through in a relatively consistent fashion to the generated text, and the version that we tested did not show substantial value bias. Therefore, this approach can be used to generate value content that follows a widely used psychological model.

Addressing our question about potential bias in ChatGPT, we did find some evidence of liberal ideological bias though, which would be implied by greater Universalism and Self-Direction values in the responses. For some of the discriminant analyses, it appears that words within the Universalism value type dictionary category were more likely to be generated, but typically for other value type prompts that are also socially in focus. Therefore, the model seems to misattribute socially oriented values and may potentially show a slight bias towards Universalism. However, as the frequency analyses across the whole corpus suggested, this may not be due to an inherent bias within the model itself, but may reflect the higher mean of the endorsement of these values in general population samples. Hence, it may reflect the preferences of human populations rather than a bias within the model per se. Furthermore, the placement of Universalism in the MDS suggests that the use of these terms in the responses may be shifted towards a more conservative value end. In balance, there is little explicit bias in these patterns and the general trends seem to follow a general human concern with certain values.

Given that we found relatively little evidence of explicit bias, the current model may be usable for both generating relatively high fidelity value content for different research and applied purposes. One application may be the generation of value statements for corporate mission statements or for generating value-driven policy statements (Scharfbillig et al., 2021). For research purposes, it may be possible to use the model to annotate large amounts of text for value content (Gilardi et al., 2023). Alternatively, it may be possible to generate value content across different domains and to then identify relevant terms and phrases for developing a context-specific value detection tool (e.g., an updated dictionary or a fine-tuned LLM). Importantly, this training may require human supervision but could significantly cut research costs.

Moving towards more theoretically relevant questions for the value domain, one of the immediate questions is whether the value content is itself driven by linguistic features and word association probabilities that are somewhat independent from the psychological needs and motivations that are thought to underlie the psychological theorizing about values. Our supplementary analysis suggests that the issue might be more complex because the analysis of PVQ items using the dictionary suggested that linguistic patterns within the PVQ items did not replicate the overall structure and the dictionary did not perform well for identifying the value content of individual items. On the other hand, when using the value names, we reduced possibly irrelevant linguistic information that may be embedded in the survey instrument items or even the definitions, yet these very brief terms performed as well or sometimes even better than the other prompts. This suggests that there are some core concepts that can be linguistically expanded and then generate content that is psychologically coherent in terms of its value orientation. To the extent that a linguistic model is able to create theory-consistent output based on core terms, the conceptualization of values may be understood as a linguistic probability association of specific words. This is an important implication of the characteristics of ChatGPT output and requires further theorizing and investigation.

At the same time, our results also suggest that some value types or domains are less clearly linguistically represented in either the training database that underlies the LLM or in the dictionary that is used as a benchmark. For example, we found some mixing of Universalism, Benevolence and Conformity value descriptions in the generated text. To some extent this may be a feature of these value types as these types are socially oriented and concerned with interpersonal orientations towards others within large groups or societies, although with different flavors: for example conformity is about not upsetting social orders or hierarchies whereas universalism is concerned with the wellbeing of all individuals within a social system. There appears some conceptual bleeding at the linguistic and conceptual level, which will make both the generation and the detection of value content more difficult. It is important to note that without full transparency about the exact training data and tuning parameters that were used to generate a specific LLM, for example through the use of model cards (Mitchell et al., 2019), it will be impossible to come to clear conclusions about these questions.

A final observation is that some of the lowest count value types (e.g., Achievement) seemed to show relative high discrimination. It might be that less frequent terms have higher specificity and information value and therefore are more easily discriminated. This is an issue that has been noted for other lexical classification tasks (Du et al., 2022)

An important limitation is that our approach is correlational. Although dictionaries are useful benchmarks, challenges of using them may include selectivity in dictionary entries, inclusion of ambiguous words in word lists and ignoring the evolution of word meanings over time (ref historical psychology review). Therefore, the observed associations could well imply deficiencies in the dictionary or even the language-based specification of the theory, that informs the theory-driven prompts (e.g., the words chosen to measure or define value types).

At the same time, we believe that our approach is innovative and merits further attention for the validation of LLMs. There are different experimental options to test LLMs, including Turing experiments (Aher et al., 2023), text annotation tasks with comparisons to human experts (Gilardi et al., 2023) and AI psychometrics (Pellert et al., 2022) or silicon sampling (Argyle et al., 2023). Our approach combines psychological theory about different value content with a theory-driven dictionary task. The use of psychological theory allows a more careful analysis of the direction of possible shifts in responses (to discern how specific output is shifted towards specific value content). By analysis the text produced, our method stays close to the data (analysis of the text generated) without relying on additional assumptions or tasks (e.g., converting responses to numeric values that are then scored according to psychometric criteria). Our method could be easily adapted to advance earlier explorations of cross-linguistic value embeddings in pre-trained language models (Arora et al., 2022). Furthermore, the evaluation with criteria that can be linked to psychometric theory does not require human oversight or classification, reducing human biases in the assessment of the output quality.

## References

Aher, G., Arriaga, R. I., & Kalai, A. T. (2023). *Using Large Language Models to Simulate Multiple Humans and Replicate Human Subject Studies* (arXiv:2208.10264). arXiv. https://doi.org/10.48550/arXiv.2208.10264

Amozegar, M. (2021). *Tweeting in Times of Crisis: Shifting Personal Value Priorities in Corporate Communications and Impact on Consumer Engagement*. https://doi.org/10.20381/RUOR-26636

Argyle, L. P., Busby, E. C., Fulda, N., Gubler, J. R., Rytting, C., & Wingate, D. (2023). Out of One, Many: Using Language Models to Simulate Human Samples. *Political Analysis*, 1–15. https://doi.org/10.1017/pan.2023.2

Arora, A., Kaffee, L.-A., & Augenstein, I. (2022). *Probing Pre-Trained Language Models for Cross-Cultural Differences in Values*. https://doi.org/10.48550/ARXIV.2203.13722

Bardi, A., Calogero, R. M., & Mullen, B. (2008). A new archival approach to the study of values and value--behavior relations: Validation of the value lexicon. *Journal of Applied Psychology*, *93*(3), 483–497. https://doi.org/10.1037/0021-9010.93.3.483

Bender, E. M., & Koller, A. (2020). Climbing towards NLU: On Meaning, Form, and Understanding in the Age of Data. *Proceedings of the 58th Annual Meeting of the Association for Computational Linguistics*, 5185–5198. https://doi.org/10.18653/v1/2020.acl-main.463

Benoit, K., Watanabe, K., Wang, H., Nulty, P., Obeng, A., Müller, S., Matsuo, A., Lowe, W., Müller, C., & Council (ERC-2011-StG 283794-QUANTESS), E. R. (2022). *quanteda: Quantitative Analysis of Textual Data* (3.2.4). https://CRAN.R-project.org/package=quanteda

Bilsky, W., Janik, M., & Schwartz, S. H. (2011). The Structural Organization of Human Values-Evidence from Three Rounds of the European Social Survey (ESS). *Journal of Cross-Cultural Psychology*, *42*(5), 759–776. https://doi.org/10.1177/0022022110362757

Boer, D., & Fischer, R. (2013). How and when do personal values guide our attitudes and sociality? Explaining cross-cultural variability in attitude-value linkages. *Psychological Bulletin*, *139*(5), 1113–1147.

Borg, I., & Mair, P. (2022). A Note on Procrustean Fittings of Noisy Configurations. *Austrian Journal of Statistics*, *51*(4), 1–9. https://doi.org/10.17713/ajs.v51i4.1423

Bowman, S. R. (2023). Eight Things to Know about Large Language Models. *Unpublished Manuscript, New York University*.

Boyd, R. L., & Pennebaker, J. W. (2017). Language-based personality: A new approach to personality in a digital world. *Current Opinion in Behavioral Sciences*, *18*, 63–68. https://doi.org/10.1016/j.cobeha.2017.07.017

Caprara, G. V., Vecchione, M., Schwartz, S. H., Schoen, H., Bain, P. G., Silvester, J., Cieciuch, J., Pavlopoulos, V., Bianchi, G., Kirmanoglu, H., Baslevent, C., Mamali, C., Manzi, J., Katayama, M., Posnova, T.,

Tabernero, C., Torres, C., Verkasalo, M., Lönnqvist, J.-E., … Caprara, M. G. (2017). Basic Values, Ideological Self-Placement, and Voting: A Cross-Cultural Study. *Cross-Cultural Research*, *51*(4), 388–411. https://doi.org/10.1177/1069397117712194

Chan, A., Salganik, R., Markelius, A., Pang, C., Rajkumar, N., Krasheninnikov, D., Langosco, L., He, Z., Duan, Y., Carroll, M., Lin, M., Mayhew, A., Collins, K., Molamohammadi, M., Burden, J., Zhao, W., Rismani, S., Voudouris, K., Bhatt, U., … Maharaj, T. (2023). *Harms from Increasingly Agentic Algorithmic Systems* (arXiv:2302.10329). arXiv. https://doi.org/10.48550/arXiv.2302.10329

Chia, O. (2023, February 14). Civil servants to soon use ChatGPT to help with research, speech writing. *The Straits Times*. https://www.straitstimes.com/tech/civil-servants-to-soon-use-chatgpt-to-help-with-research-speech-writing

Constantz, J. (2023, March 20). Nearly Half of Firms Are Drafting Policies on ChatGPT Use. *Bloomberg.Com*. https://www.bloomberg.com/news/articles/2023-03-20/using-chatgpt-at-work-nearly-half-of-firms-are-drafting-policies-on-its-use

Du, A., Karl, J., Luczak-Roesch, M., Fetvadjiev, V., & Fischer, R. (2022). *Tracing the evolution of personality cognition in early human civilizations: A computational analysis of the Gilgamesh epic*. https://doi.org/10.31234/osf.io/f8ujw

Fischer, R. (2017). *Personality, Values, Culture: An Evolutionary Approach*. Cambridge University Press. https://doi.org/10.1017/9781316091944

Fischer, R., & Fontaine, J. R. (2010). Methods for investigating structural equivalence. *Cross-Cultural Research Methods in Psychology*, 179–215.

Fischer, R., & Karl, J. (2022). *Unravelling values and wellbeing – disentangling temporal, within- and between-person dynamics via a psychometric network perspective*. PsyArXiv. https://doi.org/10.31234/osf.io/5je2r

Fischer, R., & Karl, J. A. (2019). A Primer to (Cross-Cultural) Multi-Group Invariance Testing Possibilities in R. *Frontiers in Psychology*, *10*. https://doi.org/10.3389/fpsyg.2019.01507

Fischer, R., Karl, J., Fetvadjiev, V., Grener, A., & Luczak-Roesch, M. (2022).

Opportunities and Challenges of Extracting Values in Autobiographical Narratives. *Frontiers in Psychology*, *13*. https://doi.org/10.3389/fpsyg.2022.886455

Gilardi, F., Alizadeh, M., & Kubli, M. (2023). *ChatGPT Outperforms Crowd-Workers for Text-Annotation Tasks* (arXiv:2303.15056). arXiv. https://doi.org/10.48550/arXiv.2303.15056

Gutierrez, C. A. G. (2021). *Analyzing and Visualizing Twitter Conversations*.

Lanning, K., Wetherell, G., Warfel, E. A., & Boyd, R. L. (2021). Changing channels? A comparison of Fox and MSNBC in 2012, 2016, and 2020. *Analyses of Social Issues and Public Policy*, *21*(1), 149–174. https://doi.org/10.1111/asap.12265

Maio, G. R. (2017). *The psychology of human values*. Routledge, Taylor & Francis Group.

Mitchell, M., Wu, S., Zaldivar, A., Barnes, P., Vasserman, L., Hutchinson, B., Spitzer, E., Raji, I. D., & Gebru, T. (2019). Model Cards for Model Reporting. *Proceedings of the Conference on Fairness, Accountability, and Transparency*, 220–229. https://doi.org/10.1145/3287560.3287596

Pellert, M., Lechner, C., Wagner, C., Rammstedt, B., & Strohmaier, M. (2022). *AI Psychometrics: Using psychometric inventories to obtain psychological profiles of large language models*. PsyArXiv. https://doi.org/10.31234/osf.io/jv5dt

Ponizovskiy, V., Ardag, M., Grigoryan, L., Boyd, R., Dobewall, H., & Holtz, P. (2020). Development and Validation of the Personal Values Dictionary: A Theory–Driven Tool for Investigating References to Basic Human Values in Text. *European Journal of Personality*, *34*(5), 885–902. https://doi.org/10.1002/per.2294

Roose, K. (2023, March 30). How Should I Use A.I. Chatbots Like ChatGPT? *The New York Times*. https://www.nytimes.com/2023/03/30/technology/ai-chatbot-chatgpt-uses-work-life.html

Rozado, D. (2023). The Political Biases of ChatGPT. *Social Sciences*, *12*(3), Article 3. https://doi.org/10.3390/socsci12030148

Sagiv, L., & Schwartz, S. H. (2022). Personal Values Across Cultures. *Annual Review of Psychology*, *73*(1), 517–546. https://doi.org/10.1146/annurev-psych-020821-125100

Sagiv, L., Sverdlik, N., & Schwarz, N. (2011). To compete or to cooperate? Values' impact

on perception and action in social dilemma games. *European Journal of Social Psychology*, *41*(1), 64–77. https://doi.org/10.1002/ejsp.729

Scharfbillig, M., Smillie, L., Mair, D., Sienkiewicz, M., Keimer, J., Pinho dos Santos, R., Vinagreiro Alves, H., Vecchione, E., & Scheunemann, L. (2021). *Values and identities :a policymaker's guide.* Publications Office of the European Union. https://data.europa.eu/doi/10.2760/022780

Schwartz, S. H. (1992). Universals in the content and structure of values: Theoretical advances and empirical tests in 20 countries. In *Advances in experimental social psychology, Vol. 25* (pp. 1–65). Academic Press. https://doi.org/10.1016/S0065-2601(08)60281-6

Schwartz, S. H., & Cieciuch, J. (2022). Measuring the Refined Theory of Individual Values in 49 Cultural Groups: Psychometrics of the Revised Portrait Value Questionnaire. *Assessment*, *29*(5), 1005–1019. https://doi.org/10.1177/1073191121998760

Schwartz, S. H., Cieciuch, J., Vecchione, M., Davidov, E., Fischer, R., Beierlein, C., Ramos, A., Verkasalo, M., Lönnqvist, J.-E., Demirutku, K., Dirilen-Gumus, O., & Konty, M. (2012). Refining the theory of basic individual values. *Journal of Personality and Social Psychology*, *103*(4), 663–688.

Tatman, R. (2017). *English Word Frequency*. https://www.kaggle.com/datasets/rtatman/english-word-frequency

Tausczik, Y. R., & Pennebaker, J. W. (2010). The Psychological Meaning of Words: LIWC and Computerized Text Analysis Methods. *Journal of Language and Social Psychology*, *29*(1), 24–54. https://doi.org/10.1177/0261927X09351676

**Supplement**
**What does ChatGPT say about human values? Exploring value bias in ChatGPT using a descriptive value theory**

**Baseline results dictionary word frequency**

We first calculated the overall frequency of all terms in the dictionary in a contemporary data set of English (Tatman, 2017). The mean and median diverged somewhat (r = .81), which indicates that individual terms within the list had an undue influence on the arithmetic mean. For this reason, we used the medians in the main text. The highest medians were observed for Achievement, followed by Power and Self-Direction, the least frequent terms were associated with Hedonism values.

Table S1. Word frequencies of value dictionary terms in the English language

| Value | # terms | Mean | SD | Median | Min | Max | Range |
|-------|---------|------|-----|--------|-----|-----|-------|
| AC | 88 | 59921373.7 | 110245172 | 17914947 | 1602365 | 637134177 | 635531812 |
| BE | 95 | 40873154.14 | 88493090.42 | 7253663 | 355980 | 611054034 | 610698054 |
| CO | 129 | 31575182.67 | 63697518.22 | 9167246 | 330980 | 396975018 | 396644038 |
| HE | 97 | 21801046.65 | 51133327.07 | 3816406 | 325901 | 320105999 | 319780098 |
| PO | 102 | 37683335.66 | 55291946.38 | 17571885.5 | 1332318 | 304201237 | 302868919 |
| SD | 140 | 43365947.56 | 97958707.87 | 11874554.5 | 295233 | 1014107316 | 1013812083 |
| SE | 85 | 33705479.35 | 76425093.23 | 8353617 | 245856 | 440416431 | 440170575 |
| ST | 120 | 15477496.94 | 27267437.94 | 5385188 | 342039 | 179794224 | 179452185 |
| TR | 109 | 16261822.51 | 51317426.54 | 5326547 | 221869 | 519537222 | 519315353 |
| UN | 103 | 39201831.49 | 67881841.86 | 11453108 | 291058 | 352051342 | 351760284 |

Table S2. Counts of dictionary terms for the PVQ57-R item prompts

| Prompts | SE | TR | CO | BE | UN | SD | ST | HE | AC | PO |
|---|---|---|---|---|---|---|---|---|---|---|
| Security—Personal | **205** | 5 | 59 | 36 | 111 | 41 | 15 | 10 | 63 | 79 |
| Security—Societal | **224** | 7 | 60 | 42 | 38 | 53 | 20 | 19 | 33 | 23 |
| Tradition | 27 | **208** | 65 | 56 | 105 | 56 | 16 | 5 | 26 | 14 |
| Conformity—Interpersonal | 21 | 2 | **79** | 42 | 67 | 50 | 15 | 19 | 16 | 41 |
| Conformity—Rules | 53 | 17 | **293** | 36 | 105 | 54 | 5 | 2 | 26 | 23 |
| Benevolence—Caring | 10 | 2 | 11 | **253** | 35 | 56 | 9 | 46 | 17 | 14 |
| Benevolence—Dependability | 6 | 11 | 82 | **256** | 21 | 40 | 12 | 14 | 20 | 17 |
| Universalism—Concern | 16 | 22 | 33 | 55 | **243** | 46 | 36 | 3 | 52 | 47 |
| Universalism—Nature | 78 | 21 | 19 | 54 | **243** | 62 | 16 | 9 | 12 | 31 |
| Universalism—Tolerance | 5 | 12 | 40 | 38 | **203** | 76 | 60 | 0 | 29 | 18 |
| Self-Direction—Action | 6 | 2 | 14 | 30 | 18 | **197** | 37 | 23 | 57 | 60 |
| Self-Direction—Thought | 2 | 0 | 11 | 28 | 17 | **175** | 41 | 9 | 53 | 26 |
| Stimulation | 14 | 1 | 7 | 30 | 24 | 86 | **223** | 32 | 85 | 39 |
| Hedonism | 15 | 3 | 5 | 33 | 45 | 62 | 59 | **184** | 41 | 20 |
| Achievement | 2 | 0 | 10 | 29 | 30 | 103 | 53 | 35 | **179** | 42 |
| Power—Dominance | 13 | 7 | 37 | 44 | 44 | 66 | 17 | 24 | 51 | **169** |
| Power—Resources | 33 | 2 | 14 | 33 | 36 | 62 | 31 | 37 | 59 | **235** |
| Aggregated | | | | | | | | | | |
| SE | **429** | 12 | 119 | 78 | 149 | 94 | 35 | 29 | 96 | 102 |
| TR | 27 | **208** | 65 | 56 | 105 | 56 | 16 | 5 | 26 | 14 |
| CO | 74 | 19 | **372** | 78 | 172 | 104 | 20 | 21 | 42 | 64 |
| BE | 16 | 13 | 93 | **509** | 56 | 96 | 21 | 60 | 37 | 31 |
| UN | 99 | 55 | 92 | 147 | **689** | 184 | 112 | 12 | 93 | 96 |
| SD | 8 | 2 | 25 | 58 | 35 | **372** | 78 | 32 | 110 | 86 |
| ST | 14 | 1 | 7 | 30 | 24 | 86 | **223** | 32 | 85 | 39 |
| HE | 15 | 3 | 5 | 33 | 45 | 62 | 59 | **184** | 41 | 20 |
| AC | 2 | 0 | 10 | 29 | 30 | 103 | 53 | 35 | **179** | 42 |
| PO | 46 | 9 | 51 | 77 | 80 | 128 | 48 | 61 | 110 | **404** |

Table S3. Counts of dictionary terms for the value definition prompts

| Prompts | SE | TR | CO | BE | UN | SD | ST | HE | AC | PO |
|---|---|---|---|---|---|---|---|---|---|---|
| Security—Personal | **98** | 0 | 10 | 12 | 89 | 22 | 6 | 6 | 19 | 12 |
| Security—Societal | **88** | 1 | 5 | 17 | 44 | 19 | 7 | 8 | 13 | 11 |
| Tradition | 19 | **82** | 33 | 46 | 40 | 17 | 10 | 9 | 5 | 6 |
| Conformity—Interpersonal | 11 | 3 | **35** | 29 | 32 | 31 | 0 | 2 | 8 | 5 |
| Conformity—Rules | 34 | 4 | **135** | 11 | 65 | 15 | 2 | 0 | 8 | 6 |
| Benevolence—Caring | 8 | 3 | 7 | **24** | 62 | 29 | 8 | 1 | 11 | 12 |
| Benevolence—Dependability | 9 | 4 | 37 | **71** | 11 | 18 | 1 | 3 | 14 | 10 |
| Universalism—Concern | 4 | 4 | 24 | 11 | **145** | 17 | 9 | 3 | 16 | 39 |
| Universalism—Nature | 17 | 11 | 15 | 7 | **107** | 9 | 8 | 4 | 6 | 14 |
| Universalism—Tolerance | 1 | 5 | 6 | 14 | **106** | 17 | 19 | 1 | 15 | 12 |
| Self-Direction—Action | 6 | 2 | 18 | 9 | 26 | **88** | 8 | 1 | 18 | 6 |
| Self-Direction—Thought | 0 | 3 | 7 | 7 | 33 | **75** | 23 | 1 | 32 | 10 |
| Stimulation | 11 | 1 | 1 | 5 | 9 | 26 | **96** | 19 | 34 | 10 |
| Hedonism | 9 | 4 | 5 | 12 | 15 | 17 | 10 | **79** | 12 | 14 |
| Achievement | 3 | 0 | 31 | 3 | 61 | 19 | 5 | 13 | **69** | 24 |
| Power—Dominance | 4 | 0 | 12 | 15 | 38 | 12 | 4 | 4 | 14 | **61** |
| Power—Resources | 19 | 4 | 1 | 15 | 23 | 12 | 2 | 22 | 19 | **86** |
| **Aggregated** | | | | | | | | | | |
| SE | **186** | 1 | 15 | 29 | 133 | 41 | 13 | 14 | 32 | 23 |
| TR | 19 | **82** | 33 | 46 | 40 | 17 | 10 | 9 | 5 | 6 |
| CO | 45 | 7 | **170** | 40 | 97 | 46 | 2 | 2 | 16 | 11 |
| BE | 17 | 7 | 44 | **95** | 73 | 47 | 9 | 4 | 25 | 22 |
| UN | 22 | 20 | 45 | 32 | **358** | 43 | 36 | 8 | 37 | 65 |
| SD | 6 | 5 | 25 | 16 | 59 | **163** | 31 | 2 | 50 | 16 |
| ST | 11 | 1 | 1 | 5 | 9 | 26 | **96** | 19 | 34 | 10 |
| HE | 9 | 4 | 5 | 12 | 15 | 17 | 10 | **79** | 12 | 14 |
| AC | 3 | 0 | 31 | 3 | 61 | 19 | 5 | 13 | **69** | 24 |
| PO | 23 | 4 | 13 | 30 | 61 | 24 | 6 | 26 | 33 | **147** |

## Table S4. Counts of dictionary terms for the value name prompts

| dim | SE | TR | CO | BE | UN | SD | ST | HE | AC | PO |
|---|---|---|---|---|---|---|---|---|---|---|
| Security—Personal | **205** | 5 | 59 | 36 | 111 | 41 | 15 | 10 | 63 | 79 |
| Security—Societal | **224** | 7 | 60 | 42 | 38 | 53 | 20 | 19 | 33 | 23 |
| Tradition | 27 | **208** | 65 | 56 | 105 | 56 | 16 | 5 | 26 | 14 |
| Conformity—Interpersonal | 21 | 2 | **79** | 42 | 67 | 50 | 15 | 19 | 16 | 41 |
| Conformity—Rules | 53 | 17 | **293** | 36 | 105 | 54 | 5 | 2 | 26 | 23 |
| Benevolence—Caring | 10 | 2 | 11 | **253** | 35 | 56 | 9 | 46 | 17 | 14 |
| Benevolence—Dependability | 6 | 11 | 82 | **256** | 21 | 40 | 12 | 14 | 20 | 17 |
| Universalism—Concern | 16 | 22 | 33 | 55 | **243** | 46 | 36 | 3 | 52 | 47 |
| Universalism—Nature | 78 | 21 | 19 | 54 | **243** | 62 | 16 | 9 | 12 | 31 |
| Universalism—Tolerance | 5 | 12 | 40 | 38 | **203** | 76 | 60 | 0 | 29 | 18 |
| Self-Direction—Action | 6 | 2 | 14 | 30 | 18 | **197** | 37 | 23 | 57 | 60 |
| Self-Direction—Thought | 2 | 0 | 11 | 28 | 17 | **175** | 41 | 9 | 53 | 26 |
| Stimulation | 14 | 1 | 7 | 30 | 24 | 86 | **223** | 32 | 85 | 39 |
| Hedonism | 15 | 3 | 5 | 33 | 45 | 62 | 59 | **184** | 41 | 20 |
| Achievement | 2 | 0 | 10 | 29 | 30 | 103 | 53 | 35 | **179** | 42 |
| Power—Dominance | 13 | 7 | 37 | 44 | 44 | 66 | 17 | 24 | 51 | **169** |
| Power—Resources | 33 | 2 | 14 | 33 | 36 | 62 | 31 | 37 | 59 | **235** |
| **Aggregated** | | | | | | | | | | |
| SE | **429** | 12 | 119 | 78 | 149 | 94 | 35 | 29 | 96 | 102 |
| TR | 27 | **208** | 65 | 56 | 105 | 56 | 16 | 5 | 26 | 14 |
| CO | 74 | 19 | **372** | 78 | 172 | 104 | 20 | 21 | 42 | 64 |
| BE | 16 | 13 | 93 | **509** | 56 | 96 | 21 | 60 | 37 | 31 |
| UN | 99 | 55 | 92 | 147 | **689** | 184 | 112 | 12 | 93 | 96 |
| SD | 8 | 2 | 25 | 58 | 35 | **372** | 78 | 32 | 110 | 86 |
| ST | 14 | 1 | 7 | 30 | 24 | 86 | **223** | 32 | 85 | 39 |
| HE | 15 | 3 | 5 | 33 | 45 | 62 | 59 | **184** | 41 | 20 |
| AC | 2 | 0 | 10 | 29 | 30 | 103 | 53 | 35 | **179** | 42 |
| PO | 46 | 9 | 51 | 77 | 80 | 128 | 48 | 61 | 110 | **404** |

**Baseline result - dictionary analysis of the PVQ items**

We examined the overall frequency of terms within the PVQ, which we used for generating the prompts within GPT.

The overall number of dictionary terms across all items varied between 14 (universalism) and 2 (achievement). The relative ordering of these frequencies overlapped somewhat with the pan-universal value mean hierarchy in nationally representative samples using the Schwartz Value Survey ($r = .49$). In relation to the overall frequency of the terms in English, the correlation with the median frequency was .01, suggesting that word frequency effects had little relationship with the usage of terms in the survey instrument.

Considering hits, that is the highest frequency of dictionary terms were identified in the content matching items, we found 100% hits for self-direction, stimulation, achievement (but note that only 1 word from the dictionary was present in the achievement PVQ items), power, conformity and benevolence. For hedonism items, 5 dictionary terms were found of which 3 belonged to hedonism. The two mismatches were for the adjacent stimulation and achievement values, which could be considered theoretically appropriate. For security, we found 9 hits out of 14 value terms that were identified in the survey (64.3% success rate). The mismatches were 'protect' (Universalism values) and 'avoid' (Conformity values). Tradition values showed 3 hits out of 5 value terms that were identified in the survey (60.0% success rate). The mismatches were 'thinking' (Self-direction) and 'culture' (universalism). Universalism items showed 11 hits out of 17 values terms identified in the survey (64.7% success rate). Mismatching terms in the survey items that were coded as different value dimensions according to the dictionary were: 'weak' (power), 'opportunities' (stimulation), 'activities' (self-direction), 'defend' (security) and 'different' (stimulation).

Important points to note: Conformity-rules items had 5 hits, whereas conformity-interpersonal items only showed 1 hit. Benevolence-dependence items had 7 hits, whereas Benevolence-needs only had 4 hits. As noted above, achievement items only featured one term that was included in the achievement dictionary. These points are important when considering the ChatPGT prompts.

When running a non-metrical MDS on the count results, as can be seen in Figure S1, the theoretically expected structure was not replicated. Sparsity of the input matrix may have contributed to this poor replication.

Table S1. MDS structure based on the dictionary count in the PVQ57-R instrument

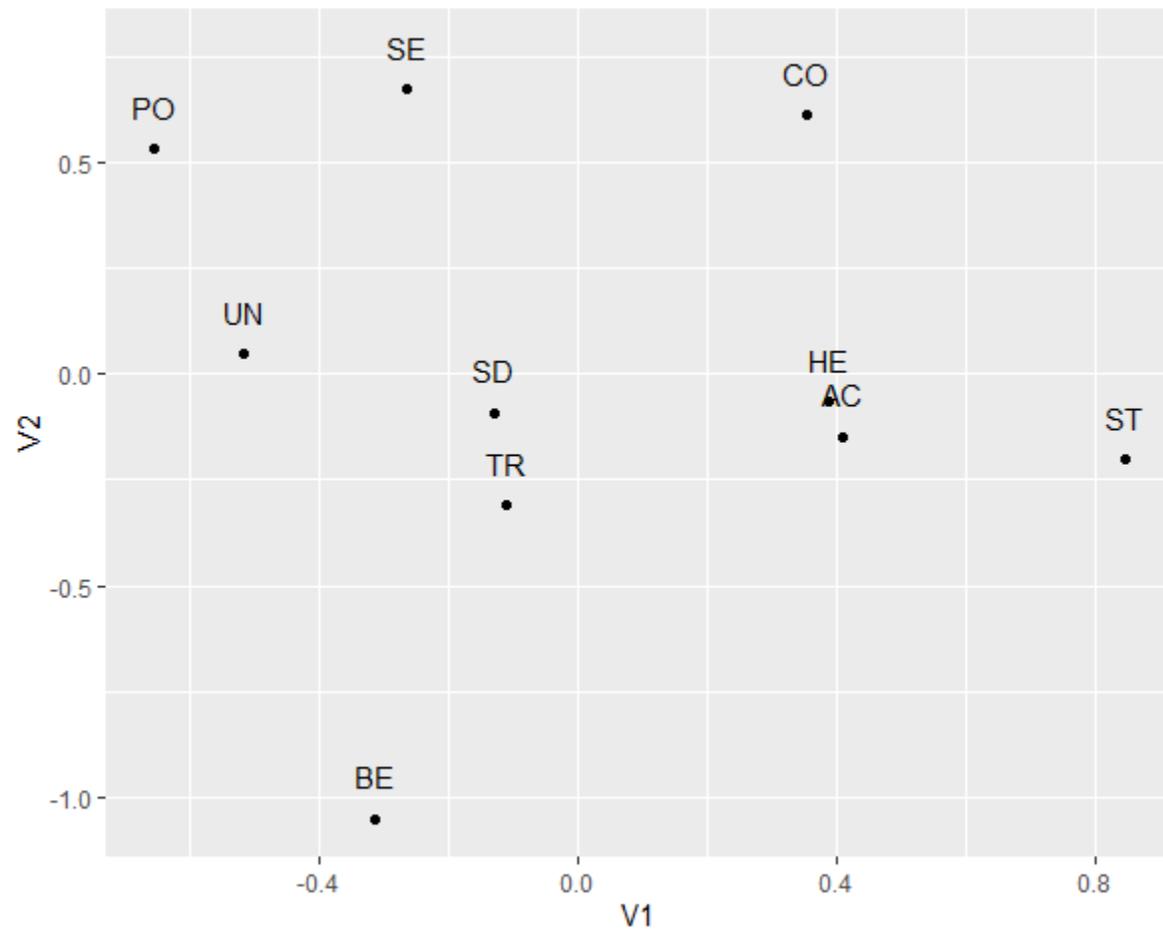